\documentclass[a4paper]{llncs}
\pdfoutput=1
\usepackage[utf8]{inputenc}
\usepackage{cwdefs}
\usepackage{amssymb}
\usepackage{amsfonts}
\usepackage{amsmath}
\usepackage{xspace}
\usepackage{fancyvrb}
\usepackage{pdfpages}
\usepackage[hidelinks]{hyperref}

\newcommand{\PIE}{\textit{PIE}\xspace}
\newcommand{\CM}{\textit{CM}\xspace}

\newcommand{\code}[1]{{\small \texttt{#1}}}

\newsavebox{\fmbox}
\newenvironment{outbox}
               {\begin{center}
                 \noindent\begin{lrbox}{\fmbox}\begin{minipage}{0.95\textwidth}}
               {\end{minipage}\end{lrbox}\fbox{\usebox{\fmbox}}
                 \end{center}}

\newcommand{\assign}{\mathrel{\mathop:}=}

\renewcommand{\f}[1]{\mathsf{#1}}
\renewcommand{\true}{\top}
\renewcommand{\false}{\bot}
\renewcommand{\imp}{\rightarrow}
\renewcommand{\revimp}{\leftarrow}
\renewcommand{\equi}{\leftrightarrow}

\newcommand{\pplmacro}[1]{\mathit{#1}}
\newcommand{\ppldefmacro}[1]{\mathit{#1}}
\newcommand{\pplparam}[1]{\mathit{#1}}

\newcommand{\pplparamplain}[1]{#1}
\newcommand{\pplparamplainidx}[2]{#1_{#2}}

\newcommand{\pplparamplainsup}[2]{#1^{#2}}

\makeatletter%
\newcommand{\entrymark}[1]{}%
\newcommand\entryhead{%
\@startsection{entry}{10}{\z@}{12pt plus 2pt minus 2pt}{0pt}{}}%
\makeatother
	    
\newcommand{\pplkbBefore}
{\entryhead*{}%
\setlength{\arraycolsep}{0pt}%
\pagebreak[0]%
\noindent%
\rule[0.5pt]{\textwidth}{2pt}\\%
\noindent}

\newcommand{\pplkbBetween}
{\setlength{\arraycolsep}{3pt}%
\\\rule[3pt]{\textwidth}{1pt}%
\par\nopagebreak\noindent Defined as\begin{center}}

\newcommand{\pplkbAfter}{\end{center}\noindent}

\newcommand{\pplkbBodyBefore}{\par\noindent where\begin{center}}
\newcommand{\pplkbBodyAfter}{\end{center}}

\newcommand{\pplIsValid}[1]{\noindent This formula is valid: $#1$\par}

\title{\PIE\ -- Proving, Interpolating and Eliminating
on the Basis of First-Order Logic\\[-10pt]}

\author{Christoph Wernhard}

\institute{Berlin, Germany\\
  \email{info@christophwernhard.com}}

\begin{document}

\maketitle

\enlargethispage{8pt}
\vspace{-15pt}
\begin{abstract}
  \PIE is a Prolog-embedded environment for automated reasoning on the basis
  of first-order logic. It includes a versatile formula macro system and
  supports the creation of documents that intersperse macro definitions,
  reasoner invocations and \LaTeX-formatted natural language text.  Invocation
  of various reasoners is supported: External provers as well as sub-systems
  of \PIE, which include preprocessors, a Prolog-based first-order prover,
  methods for Craig interpolation and methods for second-order quantifier
  elimination.
\vspace{-10pt}  
\end{abstract}

\section{Introduction}
\label{sec-intro}

First-order logic is used widely and in many roles in philosophy, mathematics,
and artificial intelligence as well as other branches of computer science.
Many practically successful reasoning approaches can be viewed as derived from
reasoning in first-order logic, for example, SAT solving, logic programming,
database query processing and reasoning in description logics.  The aim of the
\name{PIE} system is to increase practicability of actual \emph{reasoning in
  first-order logic}.  The system is written and embedded in \name{SWI-Prolog}
\cite{swiprolog}.  Formulas are basically represented by Prolog ground terms,
where explicit quantifiers distinguish Prolog atoms as formula variables.  In
addition, \PIE supports clausal formulas represented as list of lists of terms
(logic literals), with variables represented by Prolog variables.  Prolog is
very well suited as basis of a formula manipulation tool, in particular since
it supports terms as readable and printable data structures that can be
immediately used to represent logic formulas, since unification and on
occasion also backtracking is quite useful to implement formula manipulating
operations and since it offers an interpreter-based environment for
development, which is also useful to develop mechanized formalizations.  The
functionality of \PIE is provided essentially in form of a library of Prolog
predicates. It includes:
\begin{itemize}
\item Support for a Prolog-readable syntax of first-order logic formulas.
 \item Formula pretty-printing in Prolog syntax and in \LaTeX.
\item A versatile formula macro processor.
\item Support for processing documents that intersperse formula macro
  definitions, reasoner invocations and \LaTeX-formatted natural language
  text.
\item Interfaces to external first-order and propositional reasoners.
\item A built-in Prolog-based first-order theorem prover.
\item Computation of first-order Craig interpolants.
\item Second-order quantifier elimination.
\item Various formula conversion operations for use in preprocessing,
  inprocessing and output presentation.
\end{itemize}
One possibility to use the system is to develop or inspect formalizations in a
machine-supported way, similar to programming in AI languages like Prolog and
Lisp, by (re-)loading documents with macro definitions and specifications of
reasoner invocations as well as evaluating expressions interactively in the
Prolog interpreter, where output formulas might be pretty-printed.  Optionally
documents can be also rendered in \LaTeX, where macro definitions as well as
output formulas are also pretty-printed, interspersed with natural language
text in the fashion of literate programming \cite{knuth:literate}.  Craig
interpolation and second-order quantifier elimination are reasoning techniques
that compute formulas. Both are supported by \PIE on the basis of first-order
logic. For interpolation it seems that most other implementations are on the
basis of propositional logics with theory extensions and specialized for
applications in verification
\cite{benedikt:2017}.\footnote{\label{foot:implem:ipol}Craig interpolation for
  first-order logic is supported by \name{Princess}
  \cite{ruemmer:ipol:jar:2011,ruemmer:ipol:beyond:2011} and by extensions of
  \name{Vampire} \cite{vampire:interpol:2010,vampire:interpol:2012}.}  For
second-order quantifier elimination and similar operations there are several
implementations based on modal and description logics, but very few on
first-order logic.\footnote{\label{foot:implem:soqe}A Web service
  \url{http://www.mettel-prover.org/scan/} invokes an implementation of the
  SCAN algorithm \cite{scan,scan:engel}. Another system is \name{DLSForgetter}
  \url{https://personalpages.manchester.ac.uk/staff/ruba.alassaf/software.html},
  which implements the \name{DLS} algorithm \cite{dls}.  An earlier
  implementation of \name{DLS} \cite{dls:gustafsson} seems to be no longer
  available.}

The system is available as free software from its homepage
\begin{center}
  \url{http://cs.christophwernhard.com/pie}.
\end{center}
It comes with several examples whose source files as well as rendered
\LaTeX\ presentations can also be accessed directly from the system page.
\name{Inspecting Gödel's Ontological Proof} is there an advanced application
demo that utilizes some of the recently introduced system features.
An earlier version of the system was presented at the 2016 workshop
\name{Practical Aspects of Automated Reasoning} \cite{cw-paar}.  Many small
improvements make it now more workable.

The rest of this system description is structured as follows: Features and
uses of the formula macro system are presented in Sect.~\ref{sec-macros}.  In
Sect.~\ref{sec-documents} the support for documents that integrate macro
definitions, reasoner invocations and natural language text is shown.
Interfaces to external provers and the prover included with \PIE are described
in Sect.~\ref{sec-provers}, followed by the discussion of included reasoners
that compute formulas by preprocessing conversions, Craig interpolation and
second-order quantifier elimination in Sect.~\ref{sec-formops}.
Section~\ref{sec-conclusion} concludes the paper.  The bibliography reflects
that the system relates to methods as well as implementation and application
aspects in a number of areas, including first-order theorem proving, Craig
interpolation, second-order quantifier elimination and knowledge
representation.

\section{Formula Macros}
\label{sec-macros}

\PIE includes a formula macro system, where macros can have parameters and
expand into first- or second-order formulas. In the simplest case, a macro
without parameters serves as a formula label that may be used in subformula
position in other formulas and is expanded into its definiens.  The following
example of such a macro definition effects that \code{kb1} is declared as
label of a formula. In the formula syntax, the comma and \code{->} represent
conjunction and implication, respectively:
\begin{center}
\begin{BVerbatim}[fontsize=\small]
def(kb1) ::
(sprinkler_was_on -> wet(grass)),
(rained_last_night -> wet(grass)),
(wet(grass) -> wet(shoes)).
\end{BVerbatim}
\end{center}
Prolog variables in macro specifications express macro parameters.  The
following example illustrates this with the schematic definition of a certain
form of abductive explanation, the \name{weakest sufficient condition}
\cite{lin:01:snc,dls-snc,cw-projcirc}, as a second-order formula, where the
parameters are the background knowledge base $\code{Kb}$, the set $\code{Na}$
(\name{non-assumables}) of predicates that are not allowed to occur free in
explanations, and the observation $\code{Ob}$.  Universal second-order
quantification is represented in the formula syntax by \code{all2}.
\begin{center}
\begin{BVerbatim}[fontsize=\small]
def(explanation(Kb, Na, Ob)) ::
all2(Na, (Kb -> Ob)).
\end{BVerbatim}
\end{center}
Pattern matching is applied to choose the effective declaration for expansion,
allowing structural recursion in macro declarations. An optional Prolog body
in a macro declaration permits expansions involving arbitrary
computations. Utility predicates for use in these bodies are provided for
common tasks. The second-order circumscription of predicate \code{P} in
formula \code{F} can thus be defined as shown in the following example, where
\code{\~} represents negation and \code{ex2} existential second-order
quantification. The suffix \code{\_p} is used for some variable names because
it is translated to the prime superscript in the \LaTeX\, rendering, as shown
below in Sect.~\ref{sec-documents}.
\begin{center}
\begin{BVerbatim}[fontsize=\small]
def(circ(P, F)) ::
F, ~ex2(P_p, (F_p, T1, ~T2)) ::-
	mac_rename_free_predicate(F, P, pn, F_p, P_p),
	mac_get_arity(P, F, A),
	mac_transfer_clauses([P/A-n], p, [P_p], T1),
	mac_transfer_clauses([P/A-n], n, [P_p], T2).
\end{BVerbatim}
\end{center}
With a macro declaration, properties of its lexical environment, in particular
configuration settings that affect the expansion, are recorded. The macro
system utilizes further features of Prolog variables to mimic some features of
the processing of lambda expressions: A Prolog variable that is free after
computing the user-specified part of the expansion is bound automatically to a
freshly generated symbol.  A Prolog variable used as macro parameter may occur
in the definiens in predicate position (\name{SWI-Prolog} has an option that
allows variables in predicate position). The parameter then can be
instantiated with a predicate symbol (Prolog atom) or a lambda term.  The
following macro definition gives an example. It specifies 2-colorability by an
existential second-order sentence and has the \name{edge} relationship as
parameter~\code{E}. The semicolon in the formula represents disjunction and
\code{all} universal first-order quantification.
\begin{center}
\begin{BVerbatim}[fontsize=\small]
def(col2(E)) ::
ex2([r,g], (all(x, (r(x) ; g(x))),
            all([x,y], (E(x,y) -> (~((r(x), r(y))),
			           ~((g(x), g(y)))))))).
\end{BVerbatim}
\end{center}
The macro can then be used with instantiating \code{E} to a predicate symbol,
as in \code{col2(e)}, or to lambda expression that might describe a particular
graph, as in \code{col2(lambda([u,v],((u=1,v=2);(u=2,v=3))))}.  At macro
expansion, \code{E(x,y)} in the definiens of \code{col2(E)} is then replaced
by \code{e(x,y)} or \code{((x=1,y=2);(x=2,y=3))}, respectively.

Macros provide a convenient way to express abstractly properties of predicates
such as transitivity and application patterns of second-order quantification
such as circumscription.  As parameterized formula labels they are helpful to
structure formalizations.
Working practically with first-order provers typically involves experimenting
with a large and developing set of related proving problems, for example with
alternate axiomatizations or different candidate theorems, and is thus often
accompanied with some meta-level technique to compose and relate the actual
proof tasks submitted to first-order reasoners. The \PIE macro system tries to
provide such a technique in a non ad-hoc, systematic way with a uniform
mechanism that remains in spirit of first-order logic, which in mathematics is
actually often used with schemas.

\section{\PIE Documents}
\label{sec-documents}

\PIE supports the handling of documents that intersperse macro definitions,
specifications of reasoning tasks and \LaTeX-formatted natural language text.
Such a \name{\PIE document} can be \emph{loaded} into the Prolog environment
like a source code file and, in addition, be \emph{processed}, which means to
invoke the specified reasoning tasks and print the \LaTeX\ fragments in the
document interspersed with \LaTeX\ presentations of the reasoning task
outputs. The resulting output \LaTeX\ document can then be displayed in PDF
format.  In such a document, the first definitions from
Sect.~\ref{sec-macros}, for example, would be rendered as follows:

\begin{outbox}
\pplkbBefore
\index{kb1@$\ppldefmacro{kb_{1}}$}$\begin{array}{lllll}
\ppldefmacro{kb_{1}}
\end{array}
$\pplkbBetween
$\begin{array}{lllll}
(\mathsf{sprinkler\_was\_on} \imp  \mathsf{wet}(\mathsf{grass})) &&&&\; \land \\
(\mathsf{rained\_last\_night} \imp  \mathsf{wet}(\mathsf{grass})) &&&&\; \land \\
(\mathsf{wet}(\mathsf{grass}) \imp  \mathsf{wet}(\mathsf{shoes})).
\end{array}
$\pplkbAfter
\end{outbox}

\begin{outbox}
\pplkbBefore
\index{explanation(Kb,Na,Ob)@$\ppldefmacro{explanation}(\pplparam{Kb},\pplparam{Na},\pplparam{Ob})$}$\begin{array}{lllll}
\ppldefmacro{explanation}(\pplparam{Kb},\pplparam{Na},\pplparam{Ob})
\end{array}
$\pplkbBetween
$\begin{array}{lllll}
\forall \pplparam{Na} \, (\pplparam{Kb} \imp  \pplparam{Ob}).
\end{array}
$\pplkbAfter
\end{outbox}

\begin{outbox}
\pplkbBefore
\index{circ(P,F)@$\ppldefmacro{circ}(\pplparamplain{P},\pplparamplain{F})$}$\begin{array}{lllll}
\ppldefmacro{circ}(\pplparamplain{P},\pplparamplain{F})
\end{array}
$\pplkbBetween
$\begin{array}{lllll}
\pplparamplain{F} \land  \lnot  \exists \pplparamplainsup{P}{\prime} \, (\pplparamplainsup{F}{\prime} \land  \pplparamplainidx{T}{1} \land  \lnot  \pplparamplainidx{T}{2}),
\end{array}
$\pplkbAfter
\pplkbBodyBefore
$
\begin{array}{l}\pplparamplainsup{F}{\prime} \assign \pplparamplain{F}[\pplparamplain{P} \mapsto \pplparamplainsup{P}{\prime}],\\
\pplparamplain{A} \assign \mathrm{arity\ of }\; \pplparamplain{P}\; \mathrm{ in }\; \pplparamplain{F},\\
\pplparamplainidx{T}{1} \assign \mathrm{transfer\ clauses}\; {[}\pplparamplain{P}/\pplparamplain{A}\textrm{-}\mathsf{n}{]} \rightarrow {[}\pplparamplainsup{P}{\prime}{]},\\
\pplparamplainidx{T}{2} \assign \mathrm{transfer\ clauses}\; {[}\pplparamplainsup{P}{\prime}{]} \rightarrow {[}\pplparamplain{P}/\pplparamplain{A}\textrm{-}\mathsf{n}{]}.

\end{array}$\pplkbBodyAfter
\end{outbox}

\noindent
First- and second-order quantifiers are not distinguished in the default
\LaTeX\ presentation.  An intuitive idea of the effect of the Prolog code in
the definition of the \code{circ} macro (presented as \name{where} clause in
\LaTeX) can be obtained by considering the expansion of an example instance:
$\pplmacro{circ}(\mathsf{p},\mathsf{p}(\mathsf{a}))$ expands into
\begin{outbox}
\[\begin{array}{lllll}
\mathsf{p}(\mathsf{a}) \land
\lnot  \exists \mathit{q} \, (\mathit{q}(\mathsf{a}) \land  \forall \mathit{x} \, (\mathit{q}(\mathit{x}) \imp  \mathsf{p}(\mathit{x})) \land  \lnot  \forall \mathit{x} \, (\mathsf{p}(\mathit{x}) \imp  \mathit{q}(\mathit{x}))).
\end{array}
\]
\end{outbox}

\noindent
\name{\PIE
  documents} serve various purposes:
\begin{itemize}
\item They are useful in the workflow of developing a formalization, in
  particular as they can be re-loaded into the Prolog environment, which
  effects appropriate updates of the specified items.

\item First-order reasoners are often heavily dependent on configuration
  settings. A \name{\PIE document} specifies all information needed to
  reproduce the results of reasoner invocations. Effective configuration
  parameters are combined by system defaults, defaults declared in the
  document and options supplied with particular specifications of reasoner
  invocations.
  
\item Formulas are presented nicely formatted in \LaTeX.  Aside of
  indentation, the \LaTeX\ pretty-printer by default applies some symbol
  conversions to subscripted or primed symbols.  Also a compact syntax where
  parentheses to separate arguments from functors and commas between arguments
  are omitted is available as an option for both Prolog and \LaTeX\ forms.

\item The interspersing of formal specifications with \LaTeX-formatted text
  allows to develop formalizations in analogy to \name{literate programming}
  \cite{knuth:literate}, that is, embedded into explanation and discussions
  formulated in natural language.
\end{itemize}

\label{sec-output-shape}
\noindent
Finding good presentations of formulas, in particular in presence of
operations that yield formulas, seems a challenging topic.  In practice often
simply conjunctive normal form is used, possibly with representing clauses as
implications. The \PIE system supports the option to present output formulas
in a shape that is similar and is obtained by computing conjunctive or
disjunctive normal form followed by un-Skolemization.

\section{Interfaced and Included Provers}
\label{sec-provers}

\PIE allows to invoke a variety of external reasoning systems: Most
first-order provers via the \name{TPTP} formats, \name{Otter}, \name{Prover9}
and \name{Mace4} via their own input format, as well as SAT and QBF solvers
via \name{DIMACS} and \name{QDIMACS}. Large propositional formulas are
supported with an internal representation based on destructive term
operations. Most features for handling propositional formulas are inherited
from the precursor system \name{ToyElim} \cite{cw-toyelim}.

A user-level predicate to test a first-order formula for validity invokes by
default first the model searcher \name{Mace4} with a short timeout, and, if it
could not find a ``counter''-model of the negated formula, the prover
\name{Prover9}, again with a short timeout.  Correspondingly, the predicate
then prints one of three result values: \name{valid}, \name{not valid} or
\name{failed to validate}.  For example, the top-level goal
\begin{center}
\begin{BVerbatim}[fontsize=\small]
:- ppl_printtime(ppl_valid((kb1, rained_last_night  -> wet(shoes)))).
\end{BVerbatim}
\end{center}
in a \PIE document effects that during document processing (at ``printtime'')
\name{Prover9} is invoked and, given the above definition of \code{kb1}, the
following is inserted into the \LaTeX\ output:

\begin{outbox}
\pplIsValid{\pplmacro{kb_{1}} \land  \mathsf{rained\_last\_night} \imp  \mathsf{wet}(\mathsf{shoes}).}
\end{outbox}

\noindent
Alternatively, the \code{ppl\_valid} statement can also be directly input as
query to the Prolog interpreter, effecting then that \code{*Valid*} is printed
to the console.  Optionally Prolog term representations of Prover9's
resolution proof or Mace4's model, respectively, can be obtained.

The \PIE system also includes the first-order Prover \CM, whose calculus can
be understood as model elimination, clausal tableau construction
\cite{handbook:tableaux:letz}, or the connection method, similar to provers of
the \name{leanCoP} family \cite{leancop,kaliszyk15:tableaux,femalecop}.  The
implementation of \CM follows the compilation-based \name{Prolog Technology
  Theorem Prover (PTTP)} paradigm \cite{pttp}. It can return clausal tableau
proofs as Prolog terms, which allow the extraction of Craig interpolants (see
Sect.~\ref{sec-craig} below).  More details and evaluation results available
at \url{http://cs.christophwernhard.com/pie/cmprover}.

\section{Beyond Proving: Operations that Output Formulas}
\label{sec-formops}

Beyond theorem proving in the strict sense and model construction, first-order
logic is related to further mechanizable operations whose results are
\emph{formulas} with certain semantic and syntactic properties and which are
supported by \PIE.

\subsection{Preprocessing Operations}
\label{sec-preprocess}

Practically successful reasoners usually apply in some way conversions of low
complexity as far as possible: as preprocessing on inputs, potentially during
reasoning, which has been termed \name{inprocessing}, and to improve the
syntactic shape of output formulas as discussed in
Sect.~\ref{sec-output-shape}.  Abstracting from these situations, we subsume
these conversions under \name{preprocessing operations}. Also the low
complexity might be taken more or less literally and, for example, be achieved
simply by trying an operation within a threshold limit of resources.  The \PIE
system includes a number of preprocessing operations including normal form
conversions, also in variants that produce structure preserving
normalizations, various simplifications of clausal formulas, and an
implementation of McCune's un-Skolemization algorithm
\cite{mccune:unskolemizing}.  While some of these preserve equivalence, others
preserve equivalence just with respect to a set of predicates, for example,
purity simplification with respect to predicates that are not deleted or
structure preserving clausification with respect to predicates that are not
added.  This can be understood as preserving the second-order equivalence
$\exists q_1 \ldots \exists q_n\, F \equiv \exists q_1 \ldots \exists q_n\,
G$, where $F$ and $G$ are in- and outputs of the conversion and $q_1, \ldots,
q_n$ are those predicates permitted to occur in $F$ or $G$ whose semantics
needs \emph{not} to be preserved.  If $q_1,\ldots, q_n$ includes all permitted
predicates, the above equivalence expresses equi-satisfiability.  Some of the
simplifications implemented in \PIE allow to specify explicitly a set of
predicates whose semantics is to be preserved, which makes them applicable for
Craig interpolation and second-order quantifier elimination discussed below.

\subsection{Craig Interpolation}
\label{sec-craig}

By Craig's interpolation theorem \cite{craig:linear}, for given first-order
formulas $F$ and $G$ such that $F$ entails $G$ a first-order formula $H$ can
be constructed such that $F$ entails $H$, $H$ entails $G$ and $H$ contains
only symbols (predicates, functions, constants, free variables) that occur in
both $F$ and $G$.  Craig interpolation has many applications in logics and
philosophy.  Main applications in computer science are in verification
\cite{mcmillan:handbook} and query reformulation, based on its relationship to
definability and construction of definientia in terms of a given vocabulary
\cite{toman:wedell:book,benedikt:book,benedikt:2017}.  The \PIE system
supports the computation of Craig interpolants~$H$ from a closed clausal
tableau that represents a proof that~$F$ entails~$G$ with a novel adaption of
Smullyan's interpolation method \cite{smullyan:book,fitting:book} for clausal
tableaux \cite{cw-ipol}.  Suitable clausal tableaux can be constructed by the
\CM prover.  \PIE also supports the conversion of proof terms returned by the
hypertableau prover \name{Hyper} \cite{cw-ekrhyper} to such tableaux and thus
to interpolants, but this is currently at an experimental
stage.\footnote{Hypertableaux, either obtained from a hypertableau prover or
  obtained from a clausal tableau prover like \CM by restructuring the tableau
  seem interesting as basis for interpolant extraction in query reformulation,
  as they allow to ensure that the interpolants are range restricted.  Some
  related preliminary results are in \cite{cw-ipol}.}  As an example, consider
a \PIE document with the top-level goal
\begin{center}
\begin{BVerbatim}[fontsize=\small]
:- ppl_printtime(ppl_ipol((all(x, p(a,x)), q) -> (ex(x, p(x,b)) ; r))).
\end{BVerbatim}
\end{center}
At document processing the interpolation procedure is invoked.  The argument
of $\code{ppl\_ipol}$ must be an implication, whose left and right side are
taken as~$F$ and~$G$, respectively. The example leads to the following
\LaTeX\, output:

\begin{outbox}
\noindent Input: $\forall \mathit{x} \, \mathsf{p}(\mathsf{a},\mathit{x}) \land  \mathsf{q} \imp  \exists \mathit{x} \, \mathsf{p}(\mathit{x},\mathsf{b}) \lor  \mathsf{r}.$\\
\noindent Result of interpolation:
\[\begin{array}{lllll}
\exists \mathit{x} \, \forall \mathit{y} \, \mathsf{p}(\mathit{x},\mathit{y}).
\end{array}
\]
\end{outbox}

\noindent
The interpolants~$H$ constructed by \PIE strengthen the requirements for Craig
interpolants in that they are actually Lyndon interpolants, that is,
predicates occur in~$H$ only in polarities in which they occur in both~$F$
and~$G$.
Symmetric interpolation
\cite[Lemma~2]{craig:uses}\cite[Sect.~5]{mcmillan:symmetric} is supported in
\PIE, implemented by computing a conventional interpolant for each of the
input formulas, corresponding to the induction suggested in \cite{craig:uses}.

In some contexts, for example the application of interpolation to compute
definientia, it is natural to use second-order formulas, although the
underlying reasoning is in fact just first-order: If $F$ and $G$ are
second-order formulas of the form of a second-order quantifier prefix that is
for $F$ just existential and for $G$ just universal, followed by a first-order
formula, then $F$ entails $G$ if and only if a first-order formula~$F'$
entails the first-order formula~$G'$, where $F'$ and $G'$ are obtained from
$F$ and $G$, respectively, by renaming quantified predicates with fresh
symbols and dropping the second-order prefixes.  An interpolant of~$F'$
and~$G'$ is then also an interpolant of $F$ and $G$. \PIE quietly translates
such second-order entailments to first-order form when submitting them to
provers.

\subsection{Second-Order Quantifier Elimination}

Second-order quantifier elimination is the task of computing for a given
formula with second-order quantifiers, that is, quantifiers upon predicate or
function symbols, an equivalent first-order formula. Of course, on the basis
of first-order logic this does not succeed in general.  Along with variants
termed \name{forgetting}, \name{uniform interpolation} or \name{projection}
its applications include deciding fragments of first-order logic, computation
of frame correspondence properties from modal axioms, computation of
circumscription, embedding nonmonotonic semantics in a classical setting,
abduction with respect to classical and to nonmonotonic semantics, forgetting
in knowledge bases, and approaches to modularization of knowledge bases
derived from the notion of conservative extension
\cite{loewenheim:15,beh:22,scan,lin-forget,dls,cw-skp,dl-conservative,soqe,grau:modular:2008,lutz:ijcai11,schmidt:2012:ackermann,cw-projcirc,cw-abduction,ks:2013:frocos,ludwig:dl14,delgrande:17}.
As already noted in the introduction, implementations of second-order
quantifier elimination on the basis of first-order logic are rare.

\PIE includes an implementation of the \name{DLS} algorithm \cite{dls} for
second-order quantifier elimination, a method based on formula rewriting until
second-order subformulas have a certain shape that allows elimination in one
step by rewriting with Ackermann's lemma, an equivalence due to
\cite{ackermann:35}.  Implementing \name{DLS} brings about many subtle issues
\cite{dls:gustafsson,dls:conradie,cw-relmon}, for example, dealing with
un-Skolemization, simplification of formulas in non-clausal form, and ensuring
success of the method for certain input classes.  The current implementation
in \PIE is far from optimum solutions of these issues, but can nevertheless be
used in nontrivial applications and might contribute to improvements by making
experiments possible.

Given the macro definitions shown in Sect.~\ref{sec-macros}
and~\ref{sec-documents}, \PIE can, for example, be used to compute abductive
explanations or circumscriptions: The top-level goal
\begin{center}
\begin{BVerbatim}[fontsize=\small]
:- ppl_printtime(ppl_elim(explanation(kb1,[wet],wet(shoes)))).
\end{BVerbatim}
\end{center}
in the document effects invocation of the elimination procedure
at document processing and printing the following phrases
in the \LaTeX\, rendering:

\begin{outbox}
\noindent Input: $\pplmacro{explanation}(\pplmacro{kb_{1}},{[}\mathsf{wet}{]},\mathsf{wet}(\mathsf{shoes})).$\\
\noindent Result of elimination:
\[\begin{array}{lllll}
\mathsf{rained\_last\_night} \lor  \mathsf{sprinkler\_was\_on}.
\end{array}
\]
\end{outbox}

\noindent
Analogously, the circumscription of $\mathsf{wet}$ in $\pplmacro{kb_{1}}$ can
be computed with the top-level goal:
\begin{center}
\begin{BVerbatim}[fontsize=\small]
:- ppl_printtime(ppl_elim(circ(wet,kb1), [simp_result=[c6]])).
\end{BVerbatim}
\end{center}
This leads to the following \LaTeX\ output:

\begin{outbox}
\noindent Input: $\pplmacro{circ}(\mathsf{wet},\pplmacro{kb_{1}}).$\\
\noindent Result of elimination:
\[\begin{array}{lllll}
(\mathsf{rained\_last\_night} \imp  \mathsf{wet}(\mathsf{grass})) &&&&\; \land \\
(\mathsf{sprinkler\_was\_on} \imp  \mathsf{wet}(\mathsf{grass})) &&&&\; \land \\
(\mathsf{wet}(\mathsf{grass}) \imp  \mathsf{wet}(\mathsf{shoes})) &&&&\; \land \\
\forall \mathit{x} \, (\mathsf{wet}(\mathit{x}) \imp  \mathsf{rained\_last\_night} \lor  \mathsf{sprinkler\_was\_on}) &&&&\; \land \\
\forall \mathit{x} \, (\mathsf{wet}(\mathit{x}) \land  \mathsf{wet}(\mathsf{grass}) \imp  \mathit{x}=\mathsf{grass} \lor  \mathit{x}=\mathsf{shoes}).
\end{array}
\]
\end{outbox}

\noindent
The option \code{[simp\_result=[c6]]} supplied here to \code{ppl\_elim}
effects that the elimination result is postprocessed by equivalence preserving
conversions with the aim to make it more readable, as discussed above in
Sect.~\ref{sec-preprocess}.  The conversion named \code{c6} chosen for this
example converts to conjunctive normal form, applies various clausal
simplifications and then converts back to a quantified first-order formula,
involving un-Skolemization if required. That the last conjunct of the result
can be replaced by the more succinct $\forall \mathit{x} \,
(\mathsf{wet}(\mathit{x}) \imp \mathit{x}=\mathsf{grass} \lor
\mathit{x}=\mathsf{shoes})$ is, however, not detected by the current
implementation.

With options \code{[printing=false, r=\textit{Result}]} the elimination result
is not print\-ed, but bound to \textit{Result} for further processing by
Prolog. Another way to access the result is with the supplied predicate
\code{last\_ppl\_result(\textit{Result})}, which may be used to define a macro
\code{def(last\_result) :: X ::- last\_ppl\_result(X).}

\PIE also includes an implementation of second-order quantifier elimination
with respect to ground atoms by an expansion method shown in
\cite{lin-forget}, which always succeeds on the basis of first-order logic.  A
second-order quantifier is there, so-to-speak, just upon a particular instance
of a predicate.

The \name{Boolean solution problem} or \name{Boolean unification with
  predicates} is a computational task related to second-order quantifier
elimination \cite{schroeder:all,rudeanu:74,cw-boolean}.  So far, \PIE includes
experimental implementations for simple cases: Quantifier-free formulas with a
technique from \cite{eberhard} and a version for finding solutions with
respect to ground atoms, in analogy to the elimination of ground atoms.

\section{Conclusion}
\label{sec-conclusion}

The \PIE system tries to supplement what is needed to use automated
first-order proving techniques for developing and analyzing formalizations.
Its main focus is on formulas, as constituents of complex formalizations and
as results of interpolation and elimination.  Special emphasis is made on
utilizing some natural relationships between first- and second-order logics.
Key features of the system are macros, \LaTeX\ pretty-printing and integration
into the Prolog programming environment.  The system mediates between
high-level logical presentation and detailed configuration of reasoning
systems.  It shows up a number of challenging and interesting open issues for
research, for example improving practical realizations of rewriting-based
second-order quantifier elimination, strengthenings of Craig interpolation
that ensure application-relevant properties such as range restriction, and
conversion of computed formulas that are basically just semantically
characterized to comprehensible presentations.  Progress in these issues can
be directly experienced and verified with the system.

\bibliographystyle{splncs04}
\bibliography{bib_pie_01}

\end{document}